\documentclass{article}
\usepackage[latin9]{inputenc}
\usepackage{geometry}
\usepackage{float}
\usepackage{amsmath}
\usepackage{amssymb}
\usepackage{graphicx}
\usepackage[authoryear]{natbib}
\usepackage{microtype}

\makeatletter

\floatstyle{ruled}
\newfloat{algorithm}{tbp}{loa}
\providecommand{\algorithmname}{Algorithm}
\floatname{algorithm}{\protect\algorithmname}

\usepackage{caption}

\usepackage{booktabs}

\usepackage{amsfonts}
\usepackage{amsthm}
\usepackage{changes}
\usepackage{url}

\usepackage{multicol}

\usepackage[most]{tcolorbox}

\definechangesauthor[name={Panos},color=blue]{PP}

\@ifundefined{showcaptionsetup}{}{%
 \PassOptionsToPackage{caption=false}{subfig}}
\usepackage{subfig}
\date{}
\makeatother

\begin{document}
\title{The sharp, the flat and the shallow: Can weakly interacting agents
learn to escape bad minima?}
\author{Nikolas Kantas\thanks{Department of Mathematics, Imperial College London, London, United
Kingdom}, Panos Parpas\thanks{Department of Computing, Imperial College London, London, United Kingdom},
Grigorios A. Pavliotis$^{*}$}
\maketitle
\begin{abstract}
An open problem in machine learning is whether flat minima generalize
better and how to compute such minima efficiently. This is a very
challenging problem. As a first step towards understanding this question
we formalize it as an optimization problem with weakly interacting
agents. We review appropriate background material from the theory
of stochastic processes and provide insights that are relevant to
practitioners. We propose an algorithmic framework for an extended
stochastic gradient Langevin dynamics and illustrate its potential.
The paper is written as a tutorial, and presents an alternative use
of multi-agent learning. Our primary focus is on the design of algorithms
for machine learning applications; however the underlying mathematical
framework is suitable for the understanding of large scale systems
of agent based models that are popular in the social sciences, economics
and finance.
\end{abstract}

\section{Introduction\label{sec:Intro} }

It is part of the machine learning folklore that ``flat minima generalize
better". This claim is intriguing and has been investigated in many
papers starting from \citet{doi:10.1162/neco.1997.9.1.1} and more
recently in \citet{keskar2016large}, \citet{NIPS2017_7176}, \citet{DBLP:journals/corr/ChaudhariCSL16}
and \citet{NIPS2018_7875}. An even more intriguing claim that appears
in the recent machine learning literature is that, at least in practice,
variants of the Stochastic Gradient Descent (SGD) method of Robbins
\& Monro, originally proposed in the 1950s (see \citet{kushner2003stochastic,bottou2018optimization}
for a modern exposition) do indeed converge to flat minima; \citet{keskar2016large,DBLP:journals/corr/abs-1711-04623,xing2018walk}.
However, the observation that SGD typically converges to flat local
minima and that these minima generalize well lacks a rigorous justification.
Firstly, there is no widely accepted or robust definition of ``flatness".
As a result, in \citet{DBLP:journals/corr/DinhPBB17} the authors
take advantage of the usual definitions of flatness to show that through
a re-parametrization, a sharp local-minimum can be transformed into
a flat one without loosing generalization. Secondly, these claims
rely on numerical experiments, which of course hold only for a particular
data-set, hyper-parameters, and the amount of computation time used.
From a theoretical perspective, it is well known that it is NP-hard
to escape from a local minimum, SGD is only guaranteed to converge
to first-order stationary points (\citet{kushner2003stochastic}),
and even the best rigorous generalization bounds are too loose to
be of use in practice (\citet{DBLP:journals/corr/abs-1802-05296}).
Thus the question of whether SGD or any of its variants compute local
minima that generalize well is currently an open question with many
implications for the theory and practice of machine learning. In this
short paper, we review several variants of SGD and illustrate that
a system of interacting, rather than i.i.d., agents performing gradient
descent can help to smooth out sharp minima and (approximately) convexify
the loss function. The proposed ideas may be useful in answering a
variant of the question above, i.e., is there a principled and systematic
way for removing ``sharp" local minima from a given loss function
and thus provably converge to a flat minimum? We believe that answering
this question is an essential first step towards understanding generalization
in state-of-the-art machine learning models.

We consider various modifications of Stochastic Gradient Langevin
Dynamics (SGLD): 
\begin{equation}
dX_{t}=-\nabla\Phi(X_{t})dt+\sqrt{2\beta^{-1}}dB_{t},\quad X_{0}\sim\eta_{0},\label{eq:basic_sde}
\end{equation}
where $B_{t}$ is a standard $d$ dimensional Brownian motion, where
$\eta_{0}$ denotes the initial distribution. In this paper we assume
that the loss function $\Phi$ and its gradient are known. When this
is not the case then the resulting SGD can be described using Langevin
dynamics with multiplicative noise (\citet{li2017stochastic}). This
case will be studied elsewhere. We assume that $\beta$ is fixed in
time. If instead $\beta$ is gradually increased using a so called
annealing schedule, then adding noise to the normal gradient flow
can help the dynamics in \eqref{eq:basic_sde} provably converge to
a global minimum of $\Phi$ (\citet{geman1986diffusions}). Our goal
is somewhat different, we are interested in methods that can smooth
the objective function and eliminate sharp local minima. Therefore
the role of the noise term in our setting is to help the trajectory
escape local minima and saddle points. 
In addition to the direct smoothing of the objective function via
convolution with an appropriate kernel that is discussed in Section
\ref{sec:direct smoothing}, in this paper we present some additional
approaches to smoothing the loss function, using mean field limits
of (weakly) interacting SGLD and tools from multiscale analysis for
stochastic differential equations. Both approaches, multiscale analysis
and mean field averaging could be viewed as principled ways to ``smooth''
or regularize $\Phi$. The resulting smoothed loss function will have
a smaller number of local minima, and each local minimum will be more
flat. We assume that flat local minima are more desirable in machine
learning applications than sharp ones.

For simplicity we will present SGLD algorithms and their mean field
or homogenized variants mainly in continuous time. The analysis of
fully discrete algorithms, based on the continuous time dynamics that
is studied in this paper, will be presented elsewhere. The SDEs considered
here are well understood in the literature, but are rarely considered
as optimization methods. This paper can be considered as an early
contribution in this direction presented with a tutorial or short
review style. We emphasize that this paper is not meant as a thorough
review and hence we will avoid technicalities and only provide some
indicative references. The organization of the paper is as follows:
in Section \ref{sec:direct smoothing} we discuss direct smoothing
of the loss function via convolutions. In Section \ref{sec:Review}
we briefly review how interacting SDEs and mean field models can be
used as extensions to SGLD. Similarly in Section \ref{sec:Homogenization-and-averaging}
we present some tools from multiscale analysis that will be useful
for the development of the algorithms that are presented in this paper.
In Section \ref{sec:Towards-an-algorithm} we put everything together
in a single algorithm. In Section \ref{sec:Numerical-Results} we
present some early numerical results and in Section \ref{sec:Conclusions}
we provide concluding remarks and ideas for extensions. 

\section{Direct smoothing of the loss function}

\label{sec:direct smoothing}

To eliminate sharp local minima one could replace the gradient term
in the basic gradient descent algorithm with a smoother version. This
idea has its origins in global optimization using continuation methods
(see e.g. \citet{more1997global,hazan2016graduated}). In order to
eliminate these local minima one could simulate the gradient descent
dynamics of a ``smoothed'' version of the cost function instead
\begin{equation}
dX_{t}=-\nabla\Phi^{h}(X_{t})dt\label{eq:ode_h}
\end{equation}
where we denote 
\begin{equation}
\Phi^{h}\left(y\right)=(G_{h}\star\Phi)(y)=\int G_{h}(y-x)\Phi(x)dx,\label{eq:integrate phi}
\end{equation}
i.e. $\star$ denotes the convolution. A typical choice for the smoothing
kernel $G_{h}$ is the Gaussian kernel with variance $h$ 
\[
G_{h}(z)=\frac{1}{\left(2\pi h\right)^{d/2}}\exp\left(-\frac{\left\Vert z\right\Vert ^{2}}{2h}\right).
\]
The technical conditions for the above modification of the gradient
to lead to smoothing can be found in \citet{wu1996effective}, where
the method is studied (in discrete time) in more detail. Regardless
of the choice of the smoothing kernel, $\Phi^{h}$ can be interpreted
as an expectation 
\[
\Phi^{h}\left(x\right)=\int\Phi(x+y)\mu(dy),
\]
for a suitably chosen probability measure $\mu$. Furthermore, (under
appropriate conditions) 
\begin{equation}
\nabla\Phi^{h}\left(x\right)=\int\nabla\Phi(x+y)\mu(dy).\label{eq:smooth_grad}
\end{equation}
An additional point here is that when $y\sim\mu$, $\nabla\Phi(x+y)$
is an unbiased estimate of the ``biased'' gradient $\nabla\Phi^{h}$.
This implies, in particular, that there is an additional option for
randomization of the smoothed gradient.

Loosely speaking the effect of $\mu$ here is to smooth $\Phi$, a
rigorous justification of this statement appears in \citet{wu1996effective}.
The first natural questions to ask is how to design $\mu$ (or $G_{h}$)
to get the desired effect of smoothing of $\Phi$? An even more pressing
question for machine learning practitioners is to find a minimum that
is independent on any other variables hidden in notation, e.g. training
data and hyper-parameters. These are challenging questions. The conventional
approach in continuation methods is to start with a very smooth model
(using a high $h$) and gradually decrease it as the method progresses.
An additional complication is that computing the integral in \eqref{eq:integrate phi}
for the type of loss functions that appear in machine learning applications
is intractable. To address these issues, in the remainder we will
look at approaches where a smoothing measure $\mu$ does not act directly
on $\Phi$ and is constructed from the stochastic process itself.

\section{Mean-field formulations of SGLD \label{sec:Review} }

Our aim is to design variants of \eqref{eq:basic_sde} so that the
process $X_{t}$ will eventually end up closer to a minimizer of $\Phi$
and as quickly as possible. The law of the diffusion process $X_{t}$
has a smooth density with respect to the Lebesgue measure that satisfies
the Fokker-Planck (forward Kolmogorov) equation: 
\[
\partial_{t}\rho=\nabla\cdot\left(\rho\nabla\left(\beta\log\rho+\Phi\right)\right),\quad\rho(0,\cdot)=\eta_{0}(\cdot).
\]
Under appropriate assumptions on the growth of the loss function $\Phi$
at infinity, the density $\rho$ converges exponentially fast in relative
entropy (Kullback-Leibler divergence) to the unique stationary state
\[
\rho_{\infty}=\frac{1}{Z}\exp(-\beta\Phi(x)),
\]
where $Z$ denotes the normalization constant (\citet[Ch. 4]{Pavl2014}).
Finding the mode or maximizer of $\rho_{\infty}$ is equivalent finding
$x^{*}$. Here $\beta$ acts as an annealing parameter, higher values
result to $\rho_{\infty}$ putting more of its mass around lower valued
local minima of $\Phi$. For high $\beta$, the SGLD \eqref{eq:basic_sde}
reduces to a deterministic gradient descent, so it could be harder
to escape from local minima. In practice, the trade-off between convergence
speed and optimality is difficult to balance.

An alternative approach is to use interacting SGLD, as opposed to
i.i.d. copies of the Langevin dynamics~\eqref{eq:basic_sde}. If
we choose the interaction law appropriately, then the resulting coupled
system of SGLD could, in principle, explore the state space more efficiently
when the agents act in a cooperative manner (\citet[Ch. 7]{schweitzer2003}).
We will consider a system of interacting SGLD of the form 
\begin{align}
dX_{t}^{i} & =-\nabla\Phi(X_{t}^{i})dt-\big(\nabla D\star\eta_{t}^{N}\big)(X_{t}^{i})dt+\sqrt{2\beta^{-1}}dB_{t}^{i},\label{eq:interacting_sde}
\end{align}
where $i=1,\ldots,N,$ $\eta_{t}^{N}=\frac{1}{N}\sum_{i=1}^{N}\delta_{X_{t}^{i}}$,
$X_{0}^{i}\sim\eta_{0}(\cdot)$. Compared to the i.i.d. SGLD \eqref{eq:basic_sde},
the dynamics \eqref{eq:interacting_sde} uses $D(x,y)$ as an interaction
potential, which, in this paper, we will take to be convex. In particular,
we will consider the so-called Curie-Weiss interaction~\citet{dawson1983critical}
\begin{equation}
D(x,y)=\frac{\lambda}{2}\|x-y\|^{2}\label{e:inter}
\end{equation}
so that each particle experiences a linear attractive (mean reverting)
force to the empirical mean of all particles 
\[
\nabla D\star\eta_{t}^{N}(X_{t}^{i})=\lambda\left(X_{t}^{i}-\frac{1}{N}\sum_{j=1}^{N}X_{t}^{j}\right).
\]
Several questions arise: how does the density of each agent $X_{t}^{i}$
change in time? What is the invariant measure of the $N-$agent dynamics?
Can we pass to the mean field limit $N\rightarrow+\infty$, both for
the law of the $N-$agent process and for the invariant measure of
the dynamics? How does the rate of convergence of the $N$-particle
dynamics compare to that of the i.i.d. SGLD? These are questions that
have been studied extensively in the mathematical physics literature,
see for instance \citet{shiino1987dynamical,dawson1983critical,gomes2018mean}
and the references therein. In recent years it has been recognized
that interacting agents and their mean field limit can be useful for
developing global optimization algorithms (\citet{pinnau2017consensus,carrillo2018analytical}).
However, the connection between SGLD, mean field limits, consensus
formation for agent based models and the training of neural networks
is still largely unexplored.

We assume that interaction between agents is due to the interaction
potential $D$ and that all agents experience the same loss function
$\Phi$. This is in slight contrast to the growing literature on mean
field models for neural networks and supervised learning. There, typically
one takes $\Phi(x)=\frac{1}{2}\sum_{i=1}^{L}\left\Vert y_{i}-\mathcal{G}(x,\theta_{i})\right\Vert ^{2}$,
so interaction could appear from the number of layers and the parameters
$\theta$ (\citet{bengio2006convex,mei2019mean,rotskoff2018neural,sirignano2018mean,chizat2018global})
or by number of data points $L$ (see \citet{E2018} for a control
theoretic perspective). Our framework is a good abstraction of popular
machine learning algorithms, e.g. federated learning \citet{kamp2018efficient},
elastic-SGD \citet{zhang2015deep}, ensemble methods for RESNET \citet{zhang2015deep}.

The interacting SGLD~\eqref{eq:interacting_sde} is exchangeable
(\citet{dawson1983critical}), i.e. the law of the process is invariant
under arbitrary permutations of agents. Under appropriate assumptions
on the loss function, and on the initial conditions the position of
each agent converges, in the limit $N\rightarrow\infty$ to the solution
of the McKean SDE 
\begin{align*}
d\bar{X}_{t} & =-\nabla\Phi(\bar{X}_{t})dt-\nabla D\star\eta_{t}(\bar{X}_{t})dt+\sqrt{2\beta^{-1}}dB_{t},\\
\eta_{t} & =\mathcal{L}aw\left(\bar{X}_{t}\right).
\end{align*}
The density of the law of the process $\bar{X}_{t}$ is given by the
McKean-Vlasov equation: 
\begin{equation}
\partial_{t}\eta=\nabla\cdot\left(\eta\nabla\left(\beta\log\eta+\Phi+D\star\eta\right)\right),\quad\eta(0,\cdot)=\eta_{0}(\cdot).\label{e:mc-vl}
\end{equation}
The analysis of ~\eqref{eq:interacting_sde} is standard when both
$\Phi$ and $D$ are convex (e.g. see \citet{malrieu2001logarithmic}
and the references therein). When the loss function $\Phi$ is non-convex,
which is the case in most machine learning applications, the analysis
becomes more involved, see \citet{cattiaux2008probabilistic,tugaut2011captivity,tugaut2014phase,del2018uniform,durmus2018elementary}.
In particular, more than one stationary states can exist.

Stationary states of the McKean-Vlasov dynamics~\eqref{e:mc-vl}
are given by solutions of the integral equation (\citet{Tamura1984})
\begin{equation}
\tilde{\eta}\left(dx\right)=\frac{1}{\tilde{Z}}\exp\left(-\beta\left(\Phi+D\star\tilde{\eta}\right)\right)dx\label{eq:eta_mf}
\end{equation}
For small values of the product $\beta\lambda$ a unique stationary
state exists, given, for the quadratic interaction potential by 
\[
\eta_{\infty}(x)=\frac{1}{Z_{\infty}}\exp\left(-\beta(\Phi(x)+\frac{\lambda}{2}\left\Vert x\right\Vert ^{2})\right).
\]
On the other hand, for $\beta\lambda$ sufficiently large, for for
$\Phi$ non-convex, multiple stationary states exist with different
stability properties. These states are given by the formula 
\[
\eta_{\infty}^{m}(x)=\frac{1}{Z_{\infty}^{m}}\exp\left(-\beta(\Phi(x)+\frac{\lambda}{2}\left\Vert x-m\right\Vert ^{2})\right),
\]
where the parameter $m$ is determined self-consistently via the equation
$m=\int x\eta_{\infty}^{m}(x)dx.$ The number of stationary states
is related to the number of critical points of the loss function;
\citet{dawson1983critical,shiino1987dynamical,tugaut2011captivity,tugaut2014phase,gomes2018mean}.
In particular, for loss functions with many local minima, there exist
many stationary states. In the limit of infinitely many local minima,
then no SGD-type algorithm can converge. This is another manifestation
of the fact that it is necessary to smooth (quasi-convexify) the loss
function $\Phi$. It is important to note that, although the rigorous
proof of such results is valid only in the mean field limit, these
issues appear also over finite intervals and for a finite number of
agents (\citet{gomes2018mean}).

Using $\Phi+D\star\tilde{\eta}$ instead of $\Phi$ is a form or regularizing
or smoothing of the cost function. From an optimization point of view,
substituting $-\nabla\Phi(x)-\nabla D\star\eta_{t}^{N}(x)$ and using
a linear interaction for $\nabla D$ is equivalent to using an $\ell_{2}$-penalty
in the objective function for the constraint: $X_{t}^{i}=\frac{1}{N}\sum_{j=1}^{N}X_{t}^{j}$,
for each agent $i$. Therefore, for an appropriate choice of the interaction
strength $\lambda$, the objective function is approximately convex.
The idea to "encourage" each agent to follow the effective loss
function $\nabla\Phi(\frac{1}{N}\sum_{j=1}^{N}X_{t}^{j})$ has featured
recently in particle gradient algorithms (\citet{nitanda2017stochastic,chizat2018global}).

We remark that the mean field dynamics can, in principle, converge
faster to the stationary state(s) than the i.i.d. copies of SGLD.
Furthermore, the McKean-Vlasov PDE has a variational structure with
respect to the free energy functional~\citet{Villani2003} 
\begin{equation}
F(\rho)=\beta^{-1}\int\rho\ln\rho+\int \Phi\rho+\frac{\lambda}{2}\int(D\star\rho)\rho.\label{e:free-ener}
\end{equation}
This variational formulation makes clear the link between the mean
field SGLD and variational inference. The details will be presented
elsewhere.

\section{Homogenization and averaging for SGLD \label{sec:Homogenization-and-averaging}}

We return our attention to \eqref{eq:basic_sde} and seek an alternative
to smoothing the loss function $\Phi$, instead of the convolution
smoothing that was introduced in Section~\ref{sec:direct smoothing}.
In Section \ref{sec:Review} we used the empirical measure $\eta_{t}^{N}$
to smooth the potential based on empirical properties of interacting
agents. In this section, we will present briefly the approach that
was developed in \citet{chaudhari2018deep} to convert \eqref{eq:basic_sde}
into the following gradient descent algorithm: 
\begin{equation}
d\tilde{X}_{t}=-\nabla\Phi^{\beta,\gamma}(\tilde{X}_{t})dt,\quad\Phi^{\beta,\gamma}\left(x\right)=\int\Phi(x-y)\rho_{\infty}^{x}(dy),\label{eq:hom_sde_lim}
\end{equation}
where $\rho_{\infty}^{x}$ is the invariant measure of $Y_{t}$ that
appears in the limit when $\epsilon\rightarrow0$ for the following
fast/slow SDE system \begin{subequations}\label{eq:hom_sde_v2} 
\begin{align}
dX_{t} & =-\nabla\Phi(X_{t}-Y_{t})dt\label{eq:hom_sde_v2_1}\\
dY_{t} & =-\frac{1}{\epsilon}\left(\frac{1}{\gamma}Y_{t}-\nabla\Phi(X_{t}-Y_{t})\right)dt+\sqrt{\frac{2\beta^{-1}}{\epsilon}}dB_{t}\label{eq:hom_sde_v2_2}
\end{align}
\end{subequations}The parameter $\epsilon$ measures scale separation.
The limit $\epsilon\rightarrow0$ can be justified using multiscale
analysis~\citet{pavliotis2008multiscale}. Note that this is a gradient
scheme for the modified loss function $\Phi(x-\frac{y}{\epsilon})+\frac{1}{2\gamma}\left\Vert \frac{y}{\epsilon}\right\Vert ^{2}$.
In particular, $\gamma$ here acts as a regularization parameter,
precisely like the (inverse of the) interaction strength $\lambda$
in the previous section. We emphasize the similarities between~\eqref{eq:integrate phi}
and \eqref{eq:hom_sde_lim}. In particular, in the formulation presented
in this section, we can calculate the regularized loss function and
its gradient by taking the expectation with respect to an appropriate
probability measure, just like the case of the standard kernel-based
smoothing.

It is important to note that the smoothed loss function in \eqref{eq:hom_sde_lim}
can also be calculated via convolution with a Gaussian kernel: 
\begin{equation}
\Phi^{\beta,\gamma}(x)=\frac{1}{\beta}\log\Big(G_{\beta^{-1}\gamma}\star\exp(-\beta\Phi)\Big).\label{e:phi-cole-hopf}
\end{equation}
This is the Cole-Hopf formula for the solution of the viscous Hamilton-Jacobi
equation with the loss function $\Phi$ as the initial condition,
posed on the time interval $[0,\gamma]$. The larger $\gamma$ is,
the more regularized the effective potential (or relative entropy)
$\Phi^{\beta,\gamma}(x)$ is. At the level of the mean field SGLD
that was studied in the previous section, this corresponds to taking
the interaction strength to be sufficiently small so that a unique
steady state for the McKean-Vlasov dynamics exists. The connection
between smoothing at the level of the effective potential and the
absence of phase transitions for the mean field dynamics is an intriguing
problem that will be studied in future work. We mention here that
the critical noise-interaction strength at which the phase transition
(multiple stationary states exist) can be calculated by solving the
equation $\mbox{Var}_{\rho_{\infty}}(x)=(\beta\lambda)^{-1}$, where
$\rho_{\infty}$ denotes the non-interacting equilibrium distribution.

We also remark that in~\citet{chaudhari2018deep} an equivalent formulation
to~\eqref{eq:hom_sde_v2}: \begin{subequations}\label{eq:hom_sde}
\begin{align}
dX_{t} & =-\frac{1}{\gamma}(X_{t}-Y_{t})dt\label{eq:hom_sde1}\\
dY_{t} & =-\frac{1}{\epsilon}\left(\nabla\Phi(Y_{t})-\frac{1}{\gamma}(X_{t}-Y_{t})\right)dt+\sqrt{\frac{2\beta^{-1}}{\epsilon}}dB_{t}.\label{eq:hom_sde2}
\end{align}
\end{subequations} Here the regularized cost appears as $\Phi(\frac{y}{\epsilon})+\frac{1}{2\gamma}\left\Vert x-\frac{y}{\epsilon}\right\Vert ^{2}$.
This form is more convenient for the numerical implementation and
is the one that will be used in Algorithm \ref{alg:hom_mf_euler}.

\section{Towards an algorithm\label{sec:Towards-an-algorithm}}

Combining~\eqref{eq:hom_sde_lim} with \eqref{eq:interacting_sde}
we obtain: 
\begin{align}
dX_{t}^{i} & =-\frac{1}{\gamma}(X_{t}^{i}-Y_{t}^{i})dt-\lambda\left(X_{t}^{i}-\frac{1}{N}\sum_{j=1}^{N}X_{t}^{j}\right)dt\label{eq:hom_mf_sde1}\\
dY_{t}^{i} & =-\frac{1}{\epsilon}\left(\nabla\Phi(Y_{t}^{i})-\frac{1}{\gamma}(X_{t}^{i}-Y_{t}^{i})\right)dt+\sqrt{\frac{2\beta^{-1}}{\epsilon}}dW_{t}^{i}\label{eq:hom_mf_sde2}
\end{align}
We will use this scheme with a sufficiently small value of $\epsilon$,
so that we are close to the homogenized limit. The theoretical justification
of this algorithm requires the study of the joint limits $\epsilon\rightarrow0$
and $N\rightarrow+\infty$. Note that these limits do not commute,
when the loss function is nonconvex and when we are also interested
in the long time limit~\citet{gomes2018mean}. For our purposes,
it is better to first let $\epsilon\rightarrow0$ and then $N\rightarrow\infty$.
This way we first obtain a smoother $\Phi$ so that many of the local
minima have disappeared; then, mean field interactions will facilitate
exploration of the state space, without suffering from the drawback
of the existence of multiple stationary states. A more detailed study
of these issues, with particular emphasis on the optimal tuning of
the algorithm, will be presented elsewhere.

To discretize \eqref{eq:hom_mf_sde1}-\eqref{eq:hom_mf_sde2} effectively
for low $\epsilon$ we will use the heterogeneous multiscale method~\citet{weinan2005analysis}
presented in Algorithm \ref{alg:hom_mf_euler}. We do not discretize
\eqref{eq:hom_mf_sde1} directly, but its integral w.r.t the invariant
distribution of \eqref{eq:hom_mf_sde2} similar to \eqref{eq:hom_sde_lim}.
This is approximated in Algorithm \ref{alg:hom_mf_euler} using time
averaging of the faster time scale that run for $M$ intermediate
iterations. We will use two step sizes $\Delta$ and $\delta$ for
for equations \eqref{eq:hom_sde1} and \eqref{eq:hom_sde2}, respectively
and let $\mathsf{X}_{n}=X_{n\delta}$ and $\mathsf{Y}_{n,M}=Y_{n\delta}$.
For convenience, we also set $\delta M=\Delta$. In Algorithm \ref{alg:hom_mf_euler}
the parameter $m'$ plays the role of a ``burn-in" time to eliminate
transient behavior. In the remainder we set it to $1$.

\begin{algorithm}
Initialize $\mathsf{X}_{0}^{i}\sim\eta_{0}$

For $n\geq1$, for $i=1,\ldots,N$

$\bullet$ Set $\mathsf{Y}_{n,0}^{i}=\mathsf{Y}_{n-1,m'+M-1}^{i}$;
for $m=1,\ldots,M$ 
\begin{align*}
\mathsf{Y}_{n,m}^{i} & =\mathsf{Y}_{n,m-1}^{i}-\frac{\delta}{\epsilon}\left(\nabla\Phi(\mathsf{Y}_{n,m-1}^{i})-\frac{1}{\gamma}(\mathsf{X}_{n-1}^{i}-\mathsf{Y}_{n,m-1}^{i})\right)+\sqrt{\frac{2\beta^{-1}\delta}{\epsilon}}Z_{n,m}^{i};\:Z_{n,m}^{i}\sim N(0,I).
\end{align*}

$\bullet$ Compute average $\mathcal{Y}_{n}^{i}=\frac{1}{(m'+M-1)}\sum_{m=m'}^{m'+M-1}\mathsf{Y}_{n,m-1}^{i}$

$\bullet$ Update 
\begin{align*}
\mathsf{X}_{n}^{i} & =\mathsf{X}_{n-1}^{i}-\frac{1}{\gamma}(\mathsf{X}_{n-1}^{i}-\mathcal{Y}_{n}^{i})\Delta-\lambda\left(\mathsf{X}_{n-1}^{i}-\frac{1}{N}\sum_{j=1}^{N}\mathsf{X}_{n-1}^{i}\right)\Delta
\end{align*}

\caption{Discretized mean field SGLD with homogenization}
\label{alg:hom_mf_euler}
\end{algorithm}

\section{Numerical Results\label{sec:Numerical-Results}}

\subsection{Toy examples}

\begin{figure*}
\centering{}\includegraphics[height=4cm]{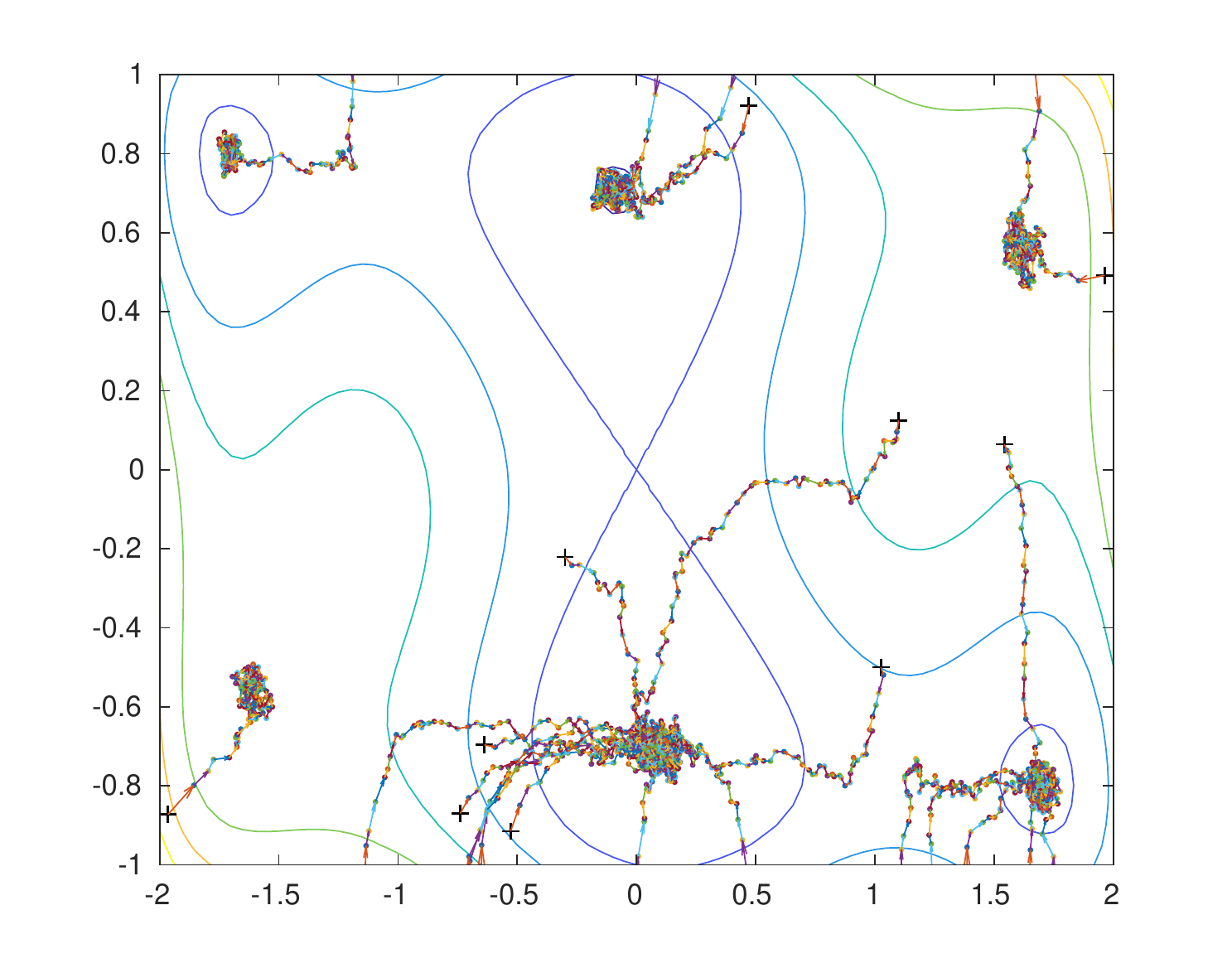} \includegraphics[height=4cm]{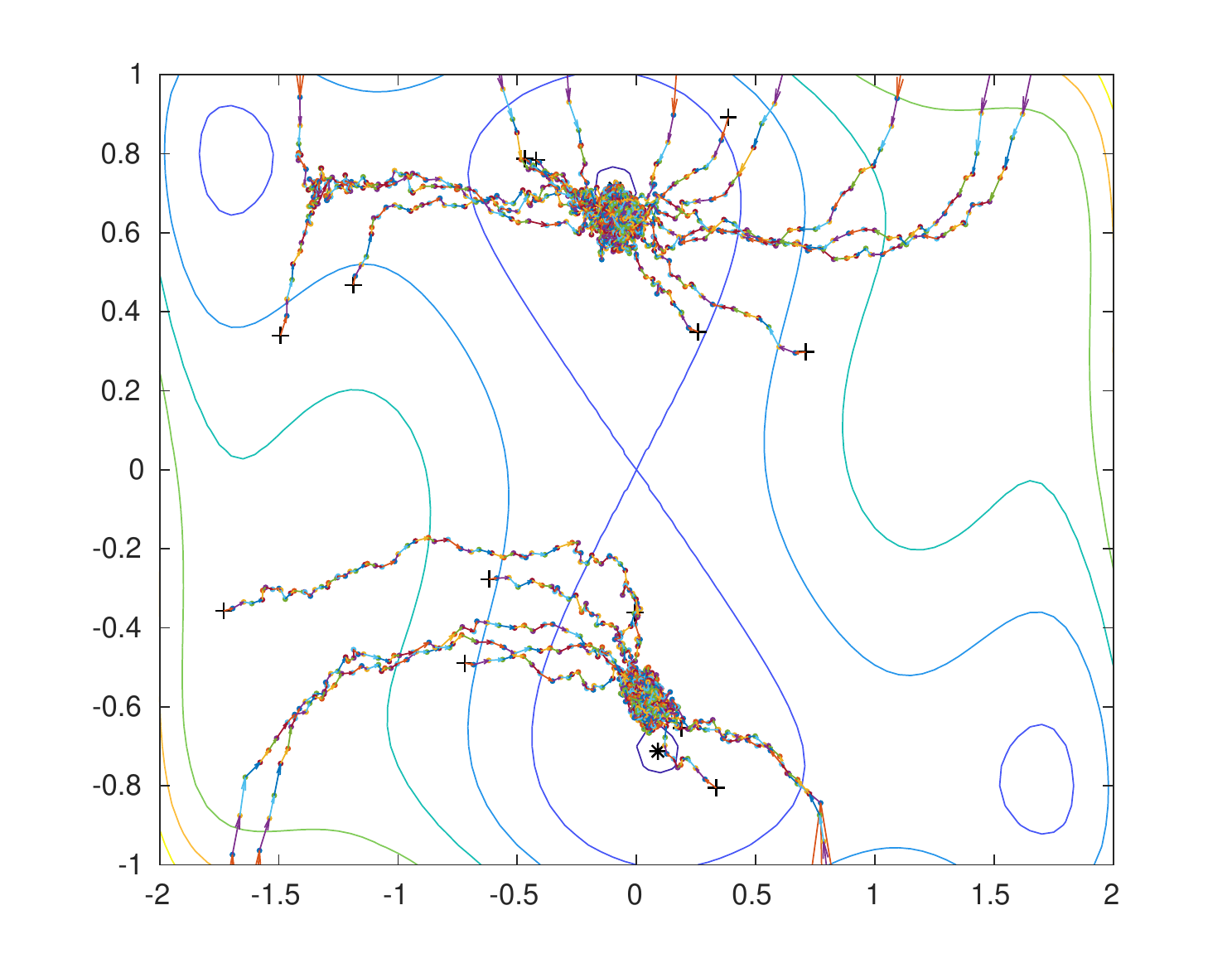}
\includegraphics[height=4cm]{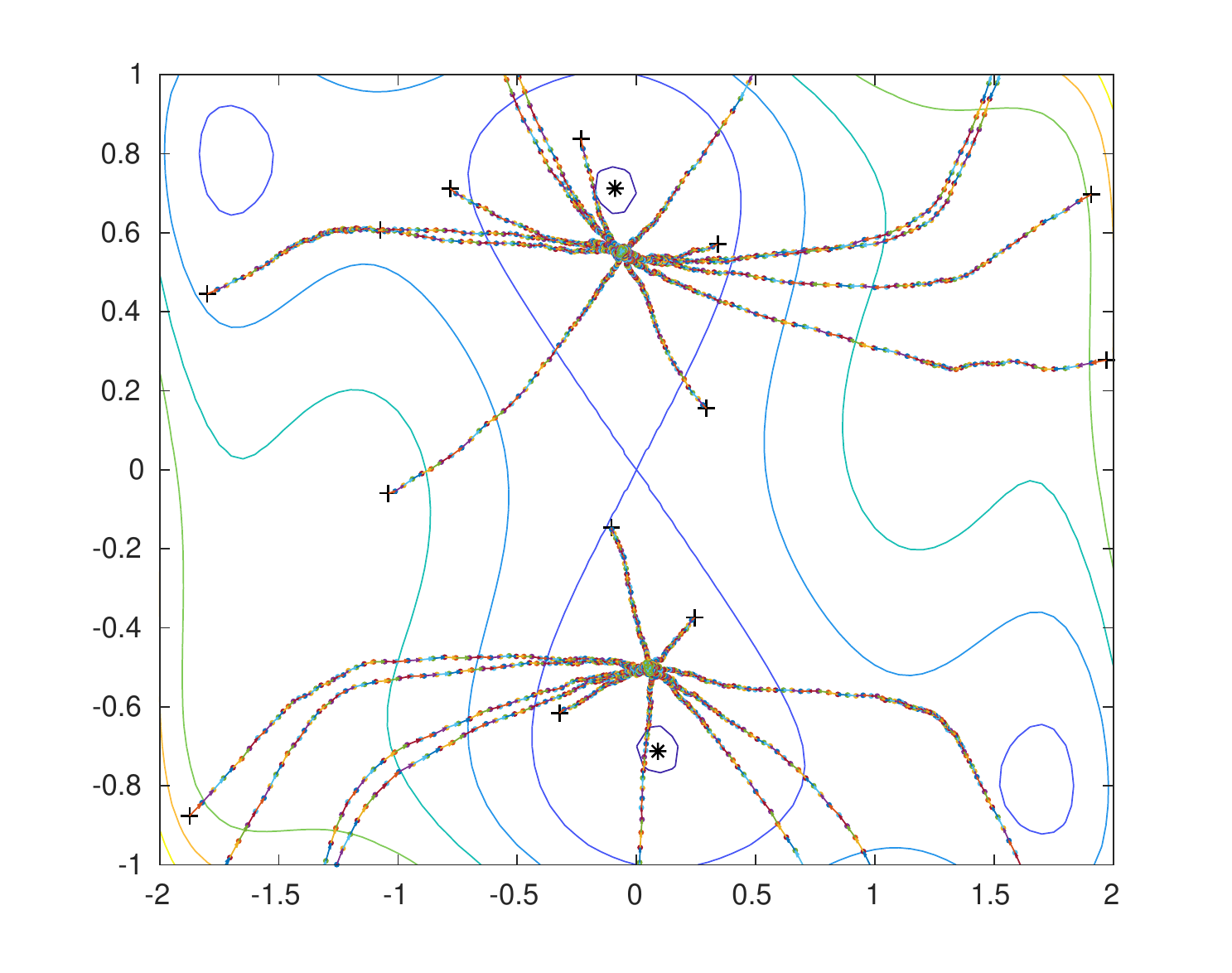} \\
 (a) i.i.d. SGLD \qquad{}\qquad{}\qquad{}(b) MF-SGD with $\lambda=2$
\qquad{}\qquad{}(c) MF-HOM-SGLD with $\lambda=2$ \\
 \includegraphics[height=4cm]{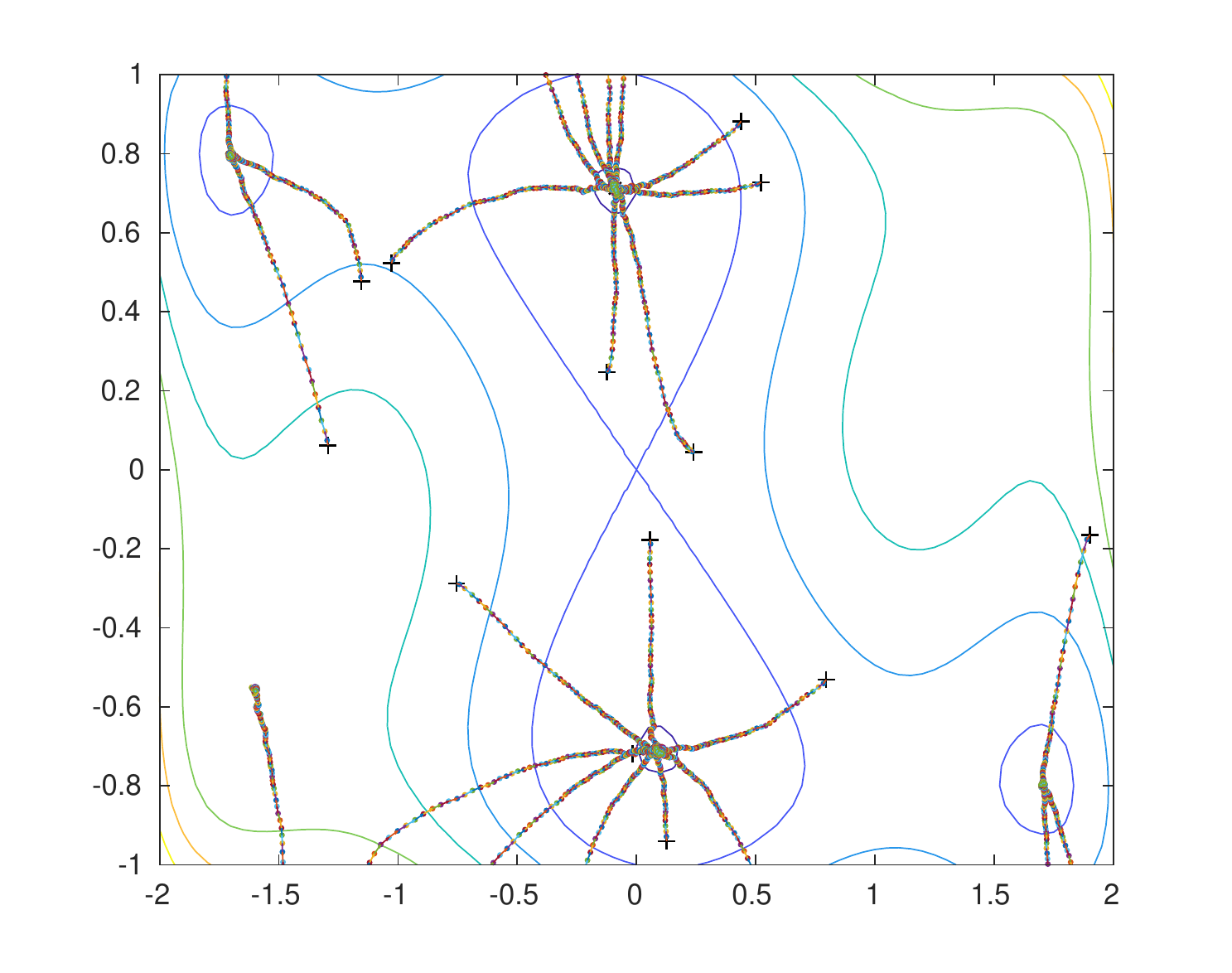} \includegraphics[height=4cm]{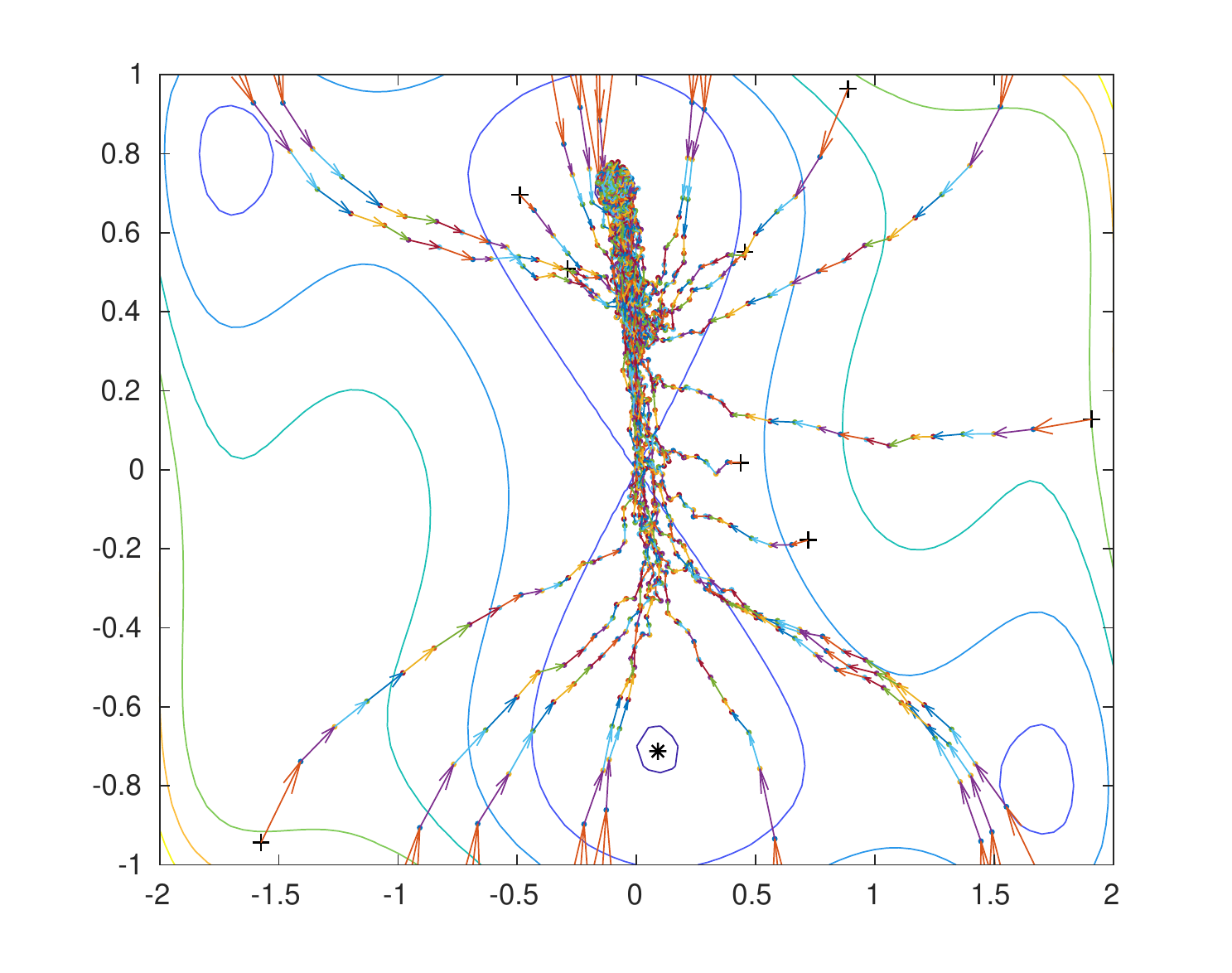}
\includegraphics[height=4cm]{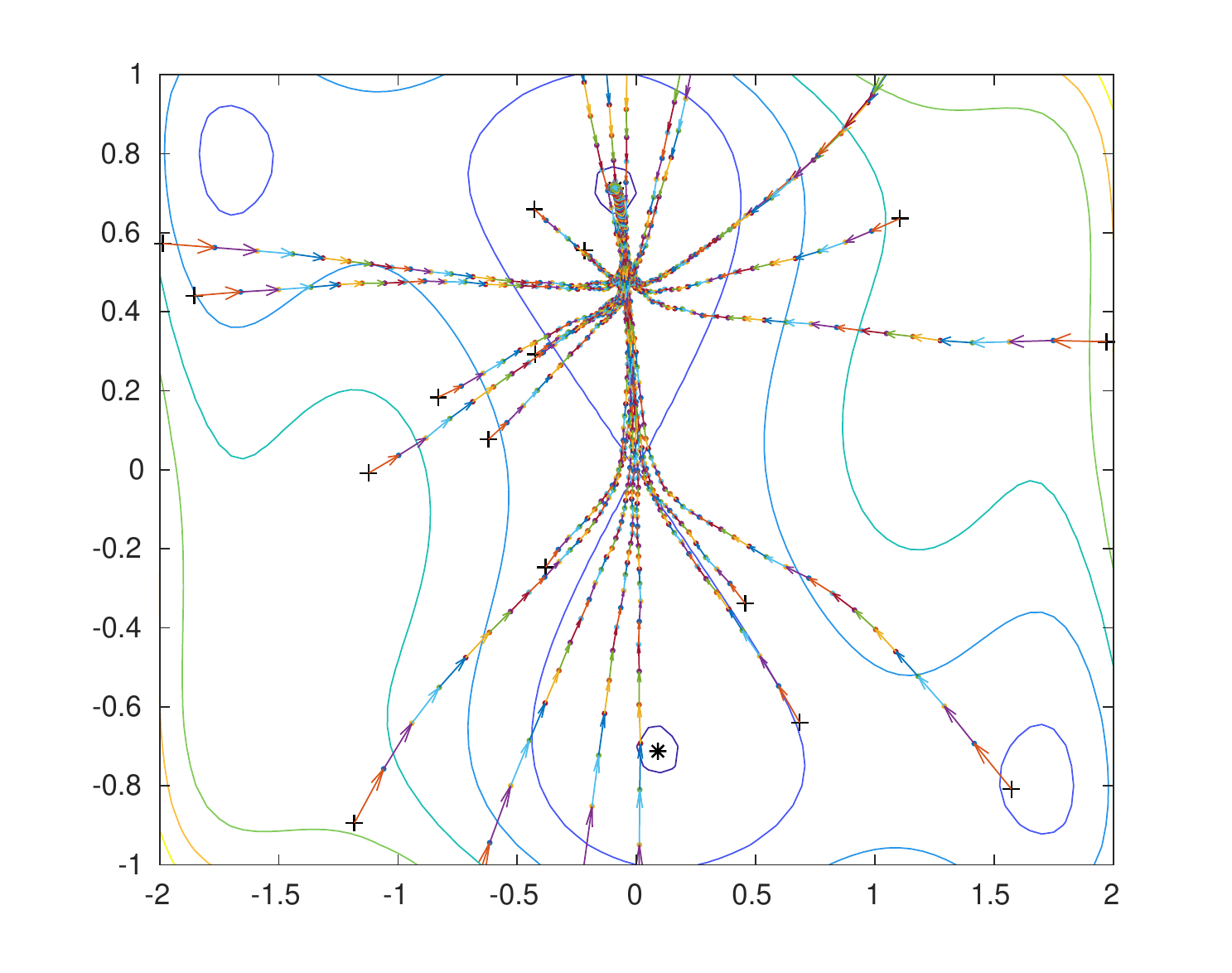} \\
 (d) i.i.d. Hom-SGD \quad{}\qquad{}\qquad{}(e) MF-SGLD with $\lambda=10$
\qquad{}\qquad{}(f) MF-HOM-SGLD with $\lambda=10$ \caption{Traces of $X_{n}^{i}$ on 2D contour plots for $\Phi$ being the six
hump camel function, $n=1,\ldots,150$. We use everywhere $N=25$,
$\beta=10$, $\eta_{0}=U[-2,2]^{2}$, $\Delta=0.01$ and for homogenization
$\epsilon=1$, $\gamma=0.1$, $m'=1$, $M=20$. Initially sampled
points are labelled by + and the two global minima by {*}.}
\label{fig:2d_visual} 
\end{figure*}

We will illustrate the performance of Algorithm \ref{alg:hom_mf_euler}
first by looking at some common problems from global optimisation.
The full algorithm will be denoted as MF-HOM-SGLD and will compare
with HOM-SGLD ($\lambda=0$), MF-SGLD (i.e. Euler discretization of
\eqref{eq:interacting_sde}) and standard SGLD. To make comparisons
fair for HOM-SGLD and SGLD we will use $N$ i.i.d copies and pick
the best performing sample/agent every time.

We begin our numerical comparisons using some two dimensional illustrations
for each method. For this we use the six hump camel function 
\[
\Phi=(4-2.1x_{1}^{2}+x_{1}^{4}/3)x_{1}^{2}+x_{1}x_{2}+(-4+4x_{2}^{2})x_{2}^{2},
\]
whose global minima are located in $(-0.0898,0.7126)$ and $(0.0898,-0.7126)$.
In Figure \ref{fig:2d_visual} we present the paths of $N=25$ agents
of each method together with contours of $\Phi$. The plots also illustrate
that homogenization smoothens the cost function gradient leading in
smoother trajectories. For the mean field methods we show paths for
a low and higher interaction case. 
Interaction strength is important in order to escape local minima
and the value for $\lambda$ influences performance directly. In this
example when $\lambda\leq1$ for both mean field methods some agents
will get stuck in local minima. As $\lambda$ increases the paths
of $X_{t}^{i}$ approach the global minima (labeled by {*} in Figure
\ref{fig:2d_visual}). The method performs well for higher values
of $\lambda$ (e.g. $\lambda=10$ in panels (e), (f)), wherby all
$X_{t}^{i}$ reach consensus and move very close to one of the global
minima. For low values (e.g. $\lambda=2$ in panels (b), (d)) $X_{t}^{i}$-s
settle at a short distance from the global minima and do not seem
to reach consensus. This is due to using an attraction towards the
empirical mean, which without consensus is close to the origin due
to the symmetry in $\Phi$. This shortcoming can be overcome using
cost weighted averages as in \citet{pinnau2017consensus}. A more
thorough investigation of such weighted interactions is left for future
work.

\begin{figure*}
\centering{}\includegraphics[width=1\textwidth]{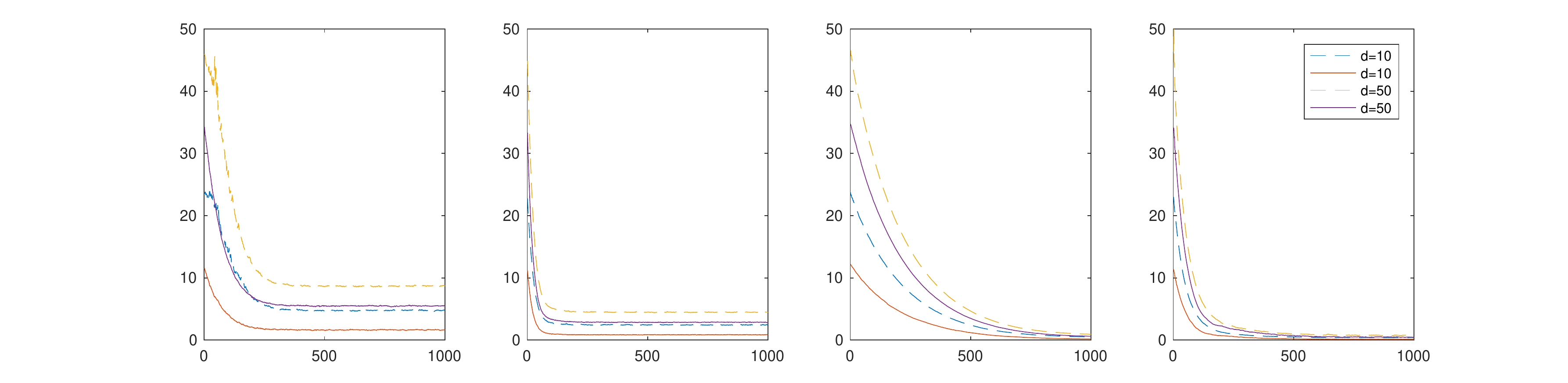}
\caption{From left to right: 
$\left\Vert X_{n}^{i^{*}}-x^{*}\right\Vert $ against $n$ for SGLD,
MF-SGLD, HOM-SGLD, MF-HOM-SGLD resp for $d=10,50$. $X^{i^{*}}$ here
is either median of worst performing agent over 50 independent runs
(dotted lines) or median of best performer (solid lines). Each run
uses $N=50$ and $\eta_{0}=U[-10,10]^{d}$}
\label{fig:x_minus_xstar}
\end{figure*}

The six hump camel function is not a very challenging problem as it
is low dimensional with a very smooth landscape and flat curvature
around each local minimum. In the context of multiple scales in $x$
and fast oscillation this was not compatible with using a low $\epsilon$.
Still homogenization performed very well. We will proceed by using
a more challenging cost function with rougher landscape and high frequency
oscillations. We will show that in this case homogenized algorithms
perform very well as smoothers of the cost. We will use $\Phi=\sum_{i=1}^{d}\left(x_{i}\sin\frac{x_{i}}{\delta}+0.1x_{i}\right)^{2}$,
where $\delta$ controls the number of local minima. 
We will consider the case where $d=10,50$, $\epsilon=\delta=0.01$.
In Figure \ref{fig:x_minus_xstar} we plot statistics from multiple
independent runs related to $\left\Vert X_{t}^{i^{*}}-x^{*}\right\Vert $
with $X_{t}^{i^{*}}$ being either the worst or best performing agent
of a given run of the algorithms. In all cases we use $\beta=1$,
$\Delta=0.01$ for MF-SGLD and SGLD and $0.005$ for the homogenized
versions, $\gamma=0.1$, $\lambda=3$, $m'=1$, $M=20$. Figure \ref{fig:x_minus_xstar}
shows a clear improvement in performance for the homogenized versions,
but this comes at the price of slower convergence. In both the plain
and homogenized SGLD the addition of mean field interaction seems
to accelerate convergence. In our simulations we noticed that Algorithm
\ref{alg:hom_mf_euler} could exhibit numerical stiffness for very
high $d$, which in practice means that a smaller step size is necessary,
so more advanced discretization schemes should be investigated.

\subsection{Co-centric Ellipse Example}

\begin{figure*}
\centering %
\noindent\begin{minipage}[c]{1\textwidth}%
 \centering \subfloat[Co-centric Ellipse Dataset]{\includegraphics[width=0.5\textwidth]{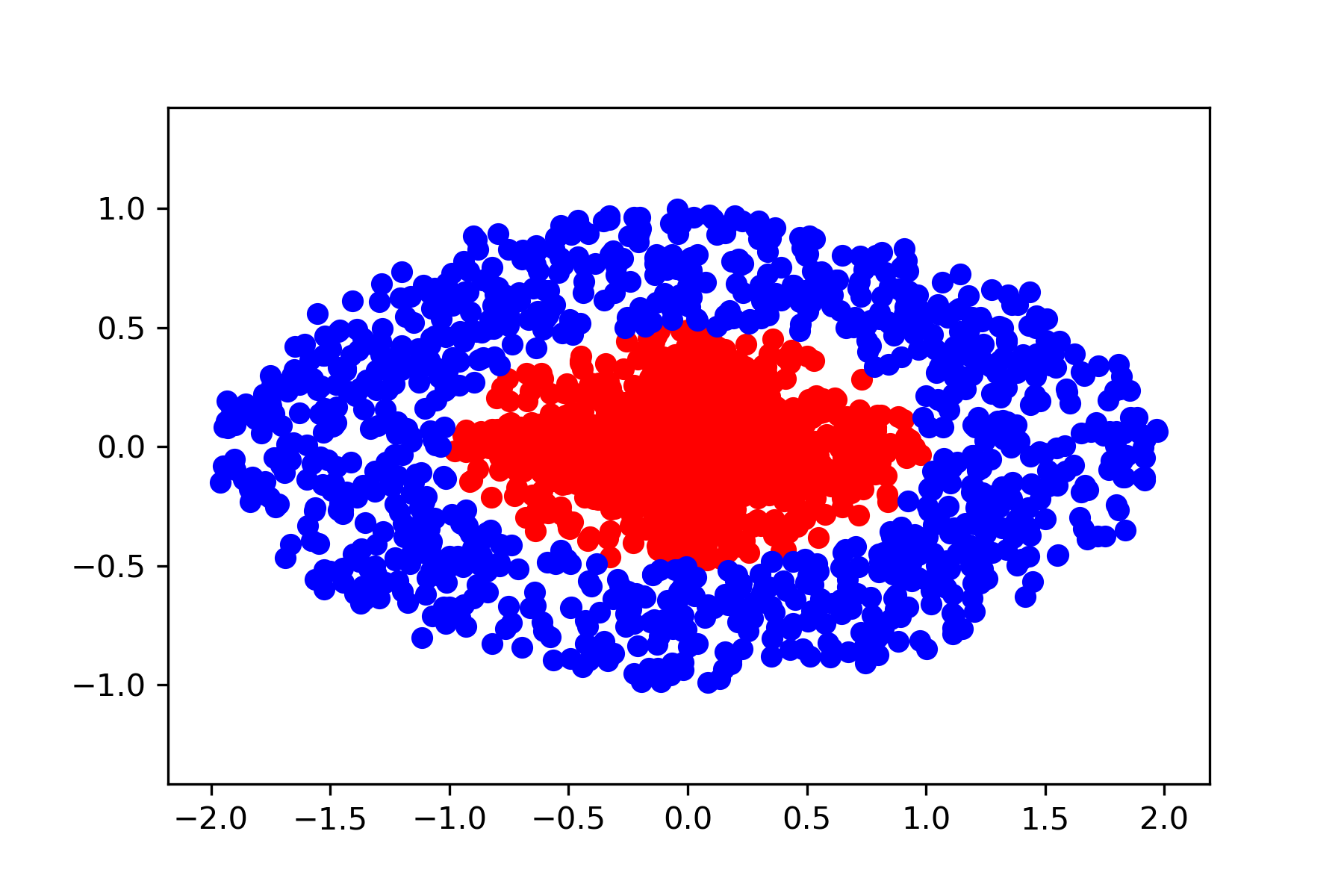}\label{fig:ellipse dataset}}\subfloat[SGLD]{\includegraphics[width=0.5\textwidth]{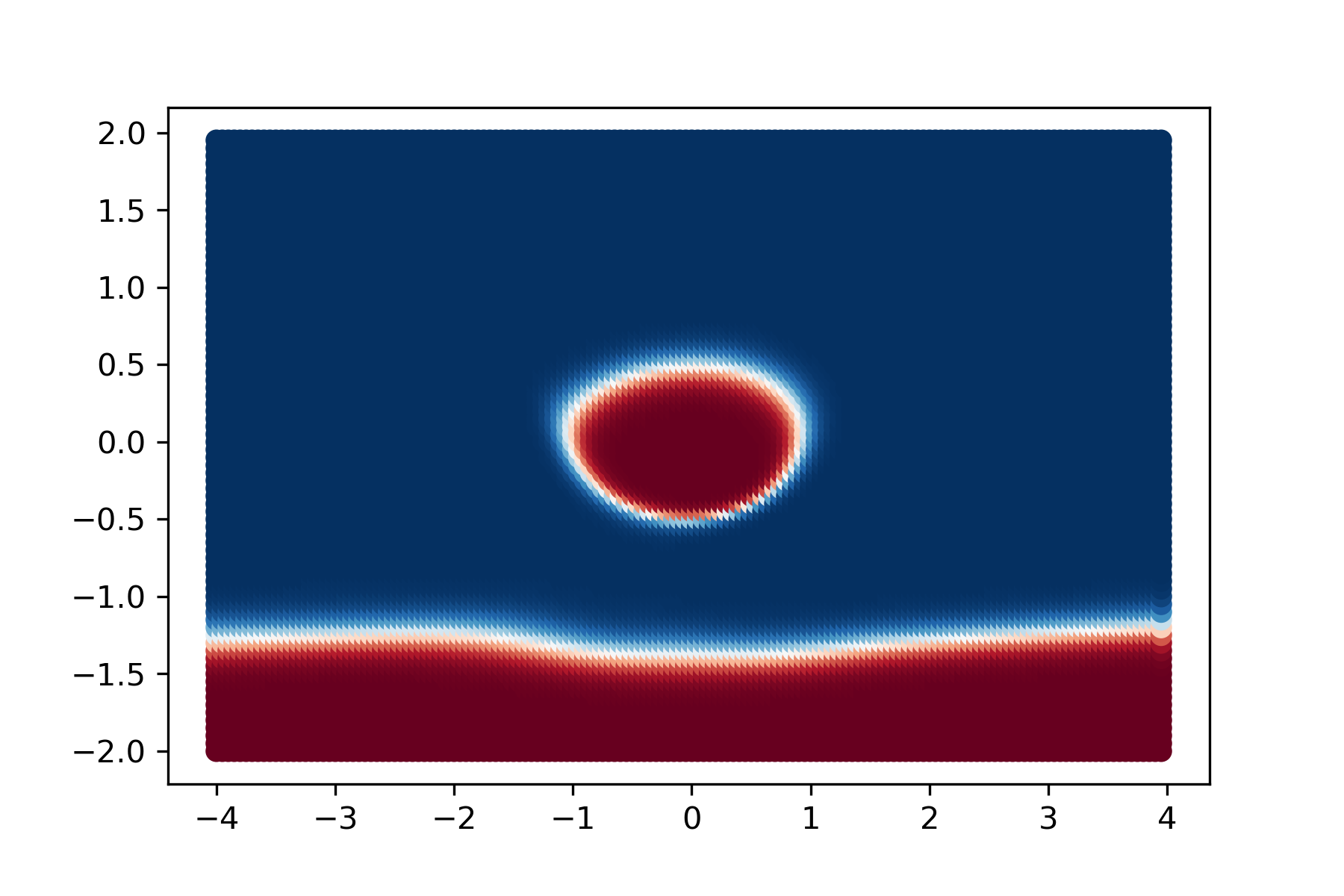}\label{fig:SGLD solution}}
\\
 \subfloat[HOM-SGLD]{\includegraphics[width=0.5\textwidth]{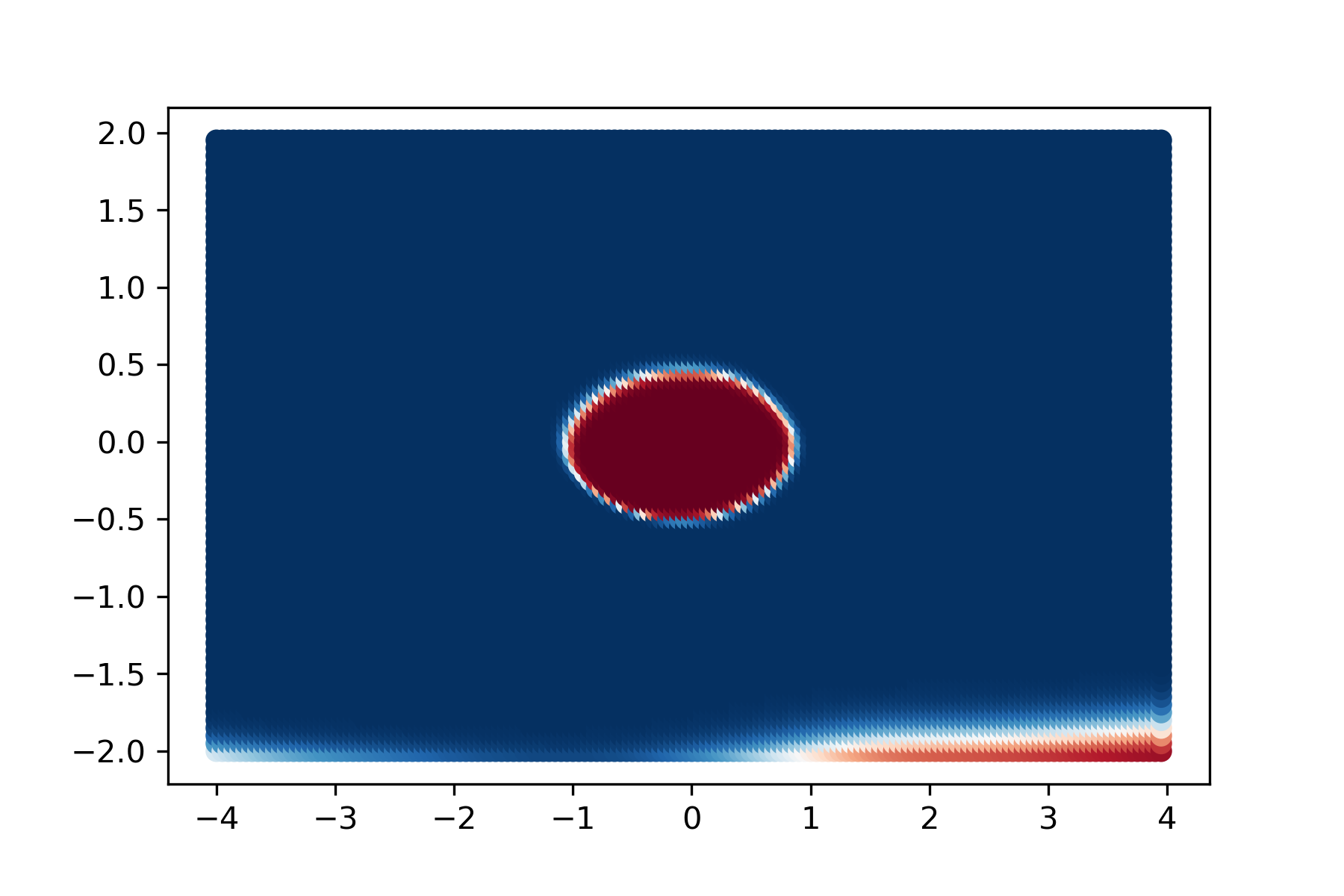}\label{fig:Regularized SGLD solution}}
\subfloat[Explicit Euler, HOM-SGLD]{\includegraphics[width=0.5\textwidth]{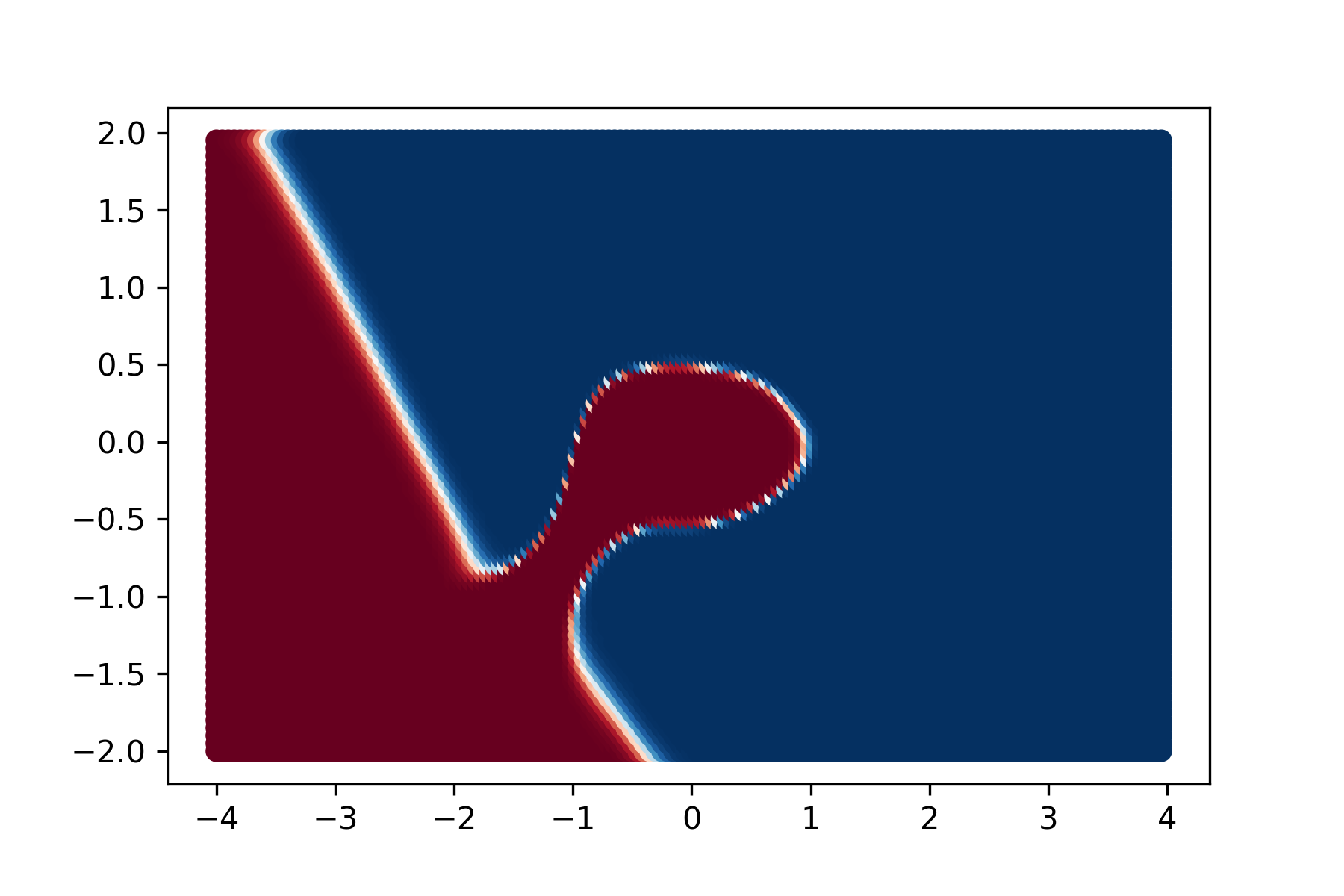}\label{fig:Euler Step}}
\caption{(a) The co-centric ellipse classification problem, Classification
probabilities with (b) SGLD and HOM-SGLD (c). (d) Failure of the explicit
Euler scheme.}
\label{fig:ellipse results} %
\end{minipage}
\end{figure*}

We describe preliminary results using the simple classification problem
of the two co-centric ellipses (see Figure \ref{fig:ellipse dataset}).
This simple classification problem is easy to solve but it neatly
illustrates the challenges of generalization. We will use this example
to demonstrate some qualitative differences between the solutions
computed by the different algorithms. We used the dynamical systems
formulation of RESNET described in \citet{0266-5611-34-1-014004}
and the serial version of the implementation described in \citet{parpas2019predict}.
After discretization with the Verlet scheme (with a step-size of 0.05),
the resulting neural network has 128 fully connected layers. We used
the $\tanh$ activation function. The resulting optimization problem
has 774 parameters and the objective function is a cross-entropy loss.
Note that the original dataset contains points in the range $[-1,1]\times[-2,2]$
(Figure \ref{fig:ellipse dataset}). We run all algorithms for 100
epochs. In Figure \ref{fig:SGLD solution} we show the classification
probability obtained from the solution obtained from SGLD. Note that
in the range $[-1,1]\times[-2,2]$ where the algorithm has observations,
the algorithm learns to classify points with the correct probability.
However, outside the original range and where the algorithm has not
seen any data points we see some strange behavior (bottom of Figure
\ref{fig:SGLD solution}). In Figure \ref{fig:Regularized SGLD solution}
we plot the results from HOM-SGLD. The solution obtained from HOM-SGLD
is valid in a broader range. Still, the solution obtained with HOM-SGLD
does eventually exhibit similar behavior to SGLD (bottom right of
Figure \ref{fig:Regularized SGLD solution}). Both algorithms reach
a similar loss function value and test accuracy, yet the solutions
behave differently outside the original data range. We conjecture
that this phenomenon will be even stronger in higher dimensions.

The example above suggests that HOM-SGLD may be more robust than SGLD,
but more research is needed to understand the behavior of all algorithms.
We illustrate this point with a slightly different numerical example.
Instead of using the Verlet scheme to obtain a discretized neural
network we use a simple explicit Euler scheme. We used the same step-size
and all the other parameters remain the same. As can be seen in Figure
\ref{fig:Euler Step} the HOM-SGLD scheme no longer performs as well
as before (other algorithms are even worse in this case). From these
experiments, we conclude that the robustness and stability of both
the underlying algorithm and model are necessary to obtain results
that can generalize.

\section{Conclusions\label{sec:Conclusions}}

We conclude by returning to the question posed in the title. In our
view the techniques described in the paper suggest that designing
algorithms using interacting agents is a very effective smoothing
strategy for loss functions with rough landscapes. While it is not
clear yet how this will benefit machine learning applications in practice,
our examples show that the method is promising.

There are many possible interesting avenues to extend the material
in this paper. For instance one could investigate different interaction
potentials $D$ such as $\ell_{1}$. Such potentials where shown in
\citet{benning2017choose} to encourage coarse changes to the dynamics
which may be desirable in non-convex optimization. In addition, there
are many aspects related to numerical analysis that could be improved.
This includes discretization methods and error analysis and possible
ways of preconditioning. For the latter parallel work in \citet{kovachki2018ensemble}
is in similar spirit and it will be interesting to see how it can
be combined with Algorithm \ref{alg:hom_mf_euler}. Finally, for the
proposed method to be useful in practice one must address the additional
computation and communication costs required to manage $N$ agents
who interact over a network.

In this paper we showed that, by considering weaklt interacting agents
and averaging, we can smooth the loss function and improve the performance
of the standard SGLD. A judicious choice of interaction between agents
can be used in order to improve the performance of algorithms in,
e.g., sampling, filtering and control applications (\citet{del2013mean}).
On the other hand, agent-based models are routinely used for the quantitative
description of phenomena in the social sciences (\citet{toscani2014}).
The understanding of the connection between these two viewpoints is
currently under investigation by our group.

\section*{Acknowledgments}
This research was funded in part by JPMorgan Chase \& Co. Any views or opinions expressed herein are solely those of the authors listed,
and may differ from the views and opinions expressed by JPMorgan Chase \& Co. or its affiliates. 
This material is not a product of the Research Department of J.P. Morgan Securities LLC. 
This material does not constitute a solicitation or offer in any jurisdiction.
The work of the authors was partly  funded by Engineering \& Physical Sciences Research Council grants EP/M028240/1, EP/P031587/1, EP/L024926/1. EP/L020564/1.

\bibliographystyle{plainnat}
\bibliography{mf.bib}

\end{document}